\documentclass[10pt,twocolumn,letterpaper]{article}

\usepackage{cvpr}              

\usepackage{hyperref}
\hypersetup{
colorlinks = true, 
urlcolor = magenta, 
linkcolor = red, 
citecolor = blue}
\usepackage{multirow}
\usepackage{enumitem}
\usepackage{romannum}
\usepackage{booktabs}
\usepackage{amsmath}
\usepackage{cuted}
\usepackage{capt-of}

\usepackage{array}
\newcolumntype{P}[1]{>{\centering\arraybackslash}p{#1}}

\def\paperID{9017} 
\def\confName{CVPR}
\def\confYear{2024}

\title{Dual Prototype Attention for Unsupervised Video Object Segmentation}
\author{Suhwan Cho$^{1,*}$\quad Minhyeok Lee$^{1,*}$\quad Seunghoon Lee$^1$\quad Dogyoon Lee$^1$\\
Heeseung Choi$^{2}$\quad Ig-Jae Kim$^{2}$\quad Sangyoun Lee$^{1}$\vspace{0.5cm}\\
$^1$~~Yonsei University\\
$^2$~~Korea Institute of Science and Technology (KIST)}

\begin{document}
\maketitle
\def\thefootnote{*}\footnotetext{These authors contribute equally to this work.}
\pagenumbering{gobble}  

\begin{abstract}
Unsupervised video object segmentation (VOS) aims to detect and segment the most salient object in videos. The primary techniques used in unsupervised VOS are 1) the collaboration of appearance and motion information; and 2) temporal fusion between different frames. This paper proposes two novel prototype-based attention mechanisms, inter-modality attention (IMA) and inter-frame attention (IFA), to incorporate these techniques via dense propagation across different modalities and frames. IMA densely integrates context information from different modalities based on a mutual refinement. IFA injects global context of a video to the query frame, enabling a full utilization of useful properties from multiple frames. Experimental results on public benchmark datasets demonstrate that our proposed approach outperforms all existing methods by a substantial margin. The proposed two components are also thoroughly validated via ablative study. Code and models are available at \url{https://github.com/Hydragon516/DPA}.
\end{abstract}

\section{Introduction}
Video object segmentation (VOS) is a fundamental task in computer vision. Given a video sequence as input, the objective is to segment objects for the entire frames. It can be divided into several categories depending on how the objects to be detected are defined. In this study, we deal with the unsupervised setting, i.e., detecting and segmenting the most salient object in a video sequence without any external guidance such as target mask or reference text.

\begin{figure}[t]
\centering
\includegraphics[width=1\linewidth]{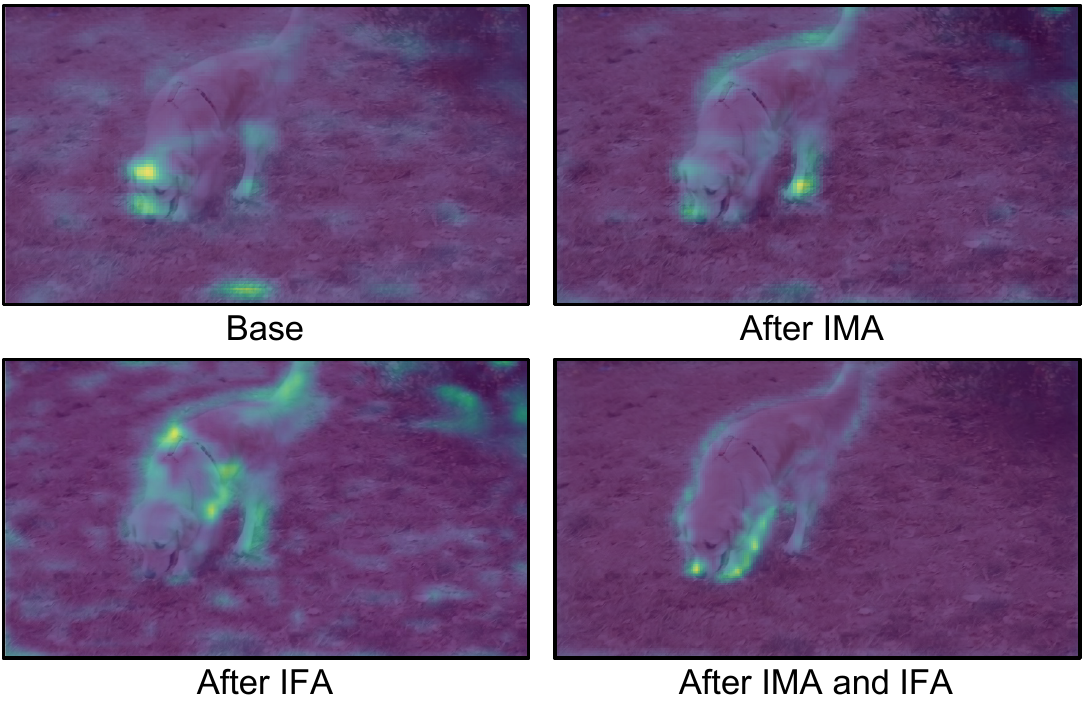}
\caption{Visualized feature maps after applying IMA and IFA.}
\label{figure1}
\end{figure}

In unsupervised VOS, collaboration of different modalities and different frames is widely adopted. As salient object usually shows distinctive movements compared to the background, existing approaches including MATNet~\cite{MATNet}, FSNet~\cite{FSNet}, and HFAN~\cite{HFAN} leverage motion cues in addition to appearance cues. For each video frame, an RGB image and an optical flow map generated by pre-trained optical flow estimation models are used as input to exploit appearance and motion information simultaneously. In order to fuse the multi-modal cues, they mainly focus on designing a reciprocity framework that blends the embedded features obtained from each modality. On the other hand, there are also studies, such as COSNet~\cite{COSNet}, AGNN~\cite{AGNN}, and AD-Net~\cite{AD-Net}, that focus on exploiting temporal coherence of a video. They transfer information of the initial frame to each query frame or iteratively refines each frame via temporally connecting different frames.

However, existing modality fusion methods and temporal aggregation methods have significant limitations. First, conventional multi-modality solutions are not carefully designed to be robust against various situations. As they fuse multi-modal cues via a direct summation, concatenation, or modulating channel weights, respective cues can act as noise if their quality is not reliable. Second, existing temporal fusion methods do not fully consider global context of a video or require high computational cost. They only consider the initial frame as an anchor frame that provides external guidance or perform iterative refinement over all frames, which severely degrades their efficiency.

In this paper, we propose two novel modules to overcome the aforementioned limitations. First, we introduce an inter-modality attention (IMA) to refine cues of respective modalities by densely integrating context information of both modalities. For each modality, useful cues are first extracted and refined to provide valuable supervision to each other. Then, instead of a naive fusion, the features of each modality is adaptively allocated to other modality based on mutual feature propagation. Second, we introduce inter-frame attention (IFA) to leverage global context of a video without requiring heavy computational cost. Once a video sequence is given as input, a designated number of frames are first sampled from the entire video sequence and features from those frames are stored in an external memory bank. When predicting each frame, the stored features are adaptively propagated to the query frames to provide overall properties of a video. Finally, we extend the proposed two modules by incorporating a prototype framework. Through converting pixel-level information to prototype-level information, more reliable and comprehensive cues can be leveraged, as each prototype is constructed as having spatial structure knowledge of the scenes.

We evaluate our proposed approach on three popular benchmark datasets, DAVIS~2016~\cite{DAVIS} validation set, FBMS~\cite{FBMS} test set, and YouTube-Objects~\cite{YTOBJ} dataset. On all of them, our method surpasses all existing methods by a substantial margin. Extensive experiments are also conducted to demonstrate the effectiveness of each proposed component.

Our main contributions can be summarized as follows:
\begin{itemize}[leftmargin=0.2in]
\item We propose dual attention modules, IMA and IFA, to effectively leverage multi-modality fusion and temporal aggregation for unsupervised VOS.

\item We incorporate a prototype framework into the proposed attention mechanisms to further improve their efficacy by refining the source information. 

\item On all public benchmark datasets, our approach sets a new state-of-the-art performance, while avoiding high computational complexity.
\end{itemize}

\begin{figure*}[t]
\centering
\includegraphics[width=1\linewidth]{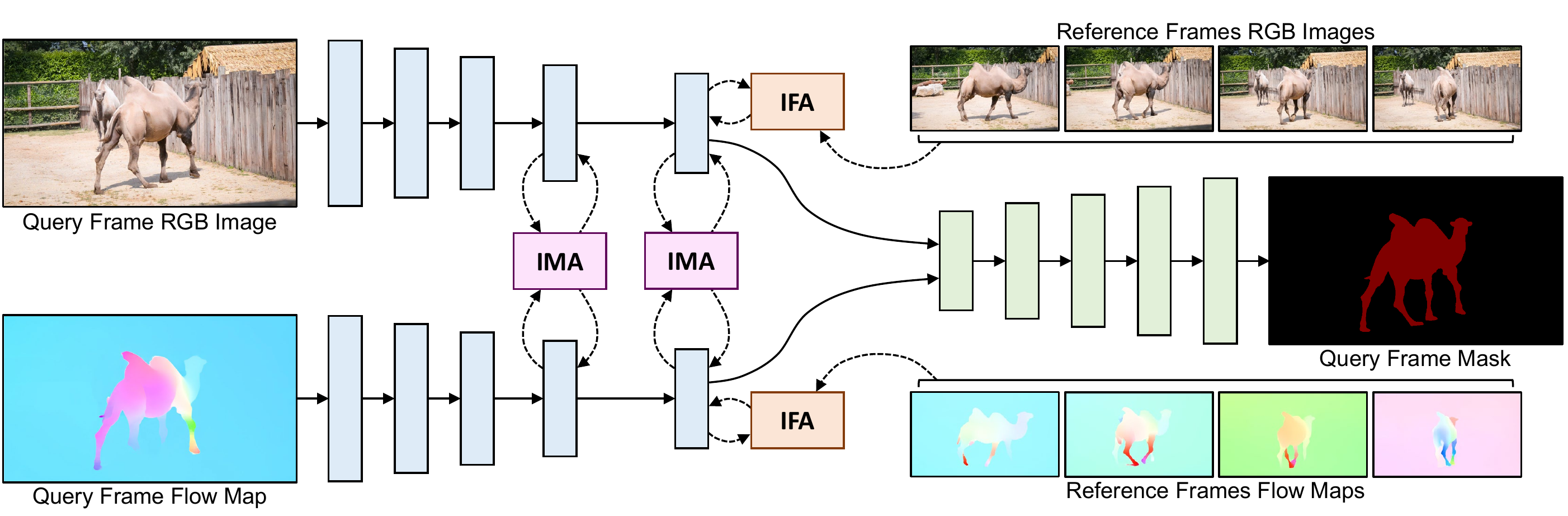}
\caption{Architecture of our proposed network. Based on a two-stream encoder-decoder architecture, IMA and IFA modules are employed. For simplicity, skip connections between encoding blocks and decoding blocks are omitted in the illustration.}
\label{figure2}
\end{figure*}

\section{Related Work}
\noindent\textbf{Multi-modality fusion.} In unsupervised VOS, two-stream architectures that jointly leverage appearance cues and motion cues have been attracting extensive attention. MATNet~\cite{MATNet} designs a two-stream encoder that utilizes an RGB image and an optical flow map to enhance spatio-temporal object representation. RTNet~\cite{RTNet} proposes a reciprocal transformation network to identify and segment primary objects in videos. FSNet~\cite{FSNet} proposes a full-duplex strategy to effectively fuse RGB images and optical flow maps; specifically, a bidirectional interaction module is used to ensure the mutual restraint between appearance and motion cues. AMC-Net~\cite{AMC-Net} proposes a co-attention gate that modulates the impacts of appearance and motion cues. Based on the learned weights, appearance and motion information can be leveraged adaptively. TransportNet~\cite{TransportNet} establishes the correspondence between appearance and motion cues while suppressing the distractions via optimal structural matching. HFAN~\cite{HFAN} proposes a hierarchical feature alignment network that aligns the object positions using the appearance and motion features. The cross-modal mismatch can be mitigated by adapting the aligned features. PMN~\cite{PMN} stores prototypes of appearance cues as well as and motion cues to fully leverage multiple modalities. TMO~\cite{TMO} optionally employs motion stream on top of appearance stream for robust learning of the motion encoder. However, the performance of these methods can be further improved as the modality fusion is performed based on a simple summation or concatenation.

\vspace{1mm}
\noindent\textbf{Temporal aggregation.} Unlike existing two-stream methods, some studies focus on fully exploiting the temporal coherence of a video. COSNet~\cite{COSNet} employs the co-attention layers to capture global correlations and scene context by propagating semantic information in the reference frames to the query frame. AGNN~\cite{AGNN} builds fully connected graphs to represent frames as nodes and relations between those frames as edges. Rich relations between arbitrary frames can be obtained through parametric message passing. AD-Net~\cite{AD-Net} and F2Net~\cite{F2Net} regard the initial frame of a video as a reference frame, and leverages the reference frame information for query frame prediction. IMP~\cite{IMP} iteratively propagates the segmentation mask of an easy reference frame to other frames by using a pre-trained semi-supervised VOS algorithm. These methods can capture temporal coherence in a video, but still suffer from certain problems such as the global context of a video not being completely leveraged and requiring heavy computational complexity owing to the iterative inferring process.

\section{Approach}
\subsection{Problem Formulation}
The goal of an unsupervised VOS algorithm is to identify the most salient object for all frames of a video. Following common protocol in the VOS community, we collaboratively use RGB images and optical flow maps as the input of our network. The network output is binary segmentation masks that have the same resolution as the input information. RGB images, optical flow maps, and output segmentation masks are denoted as $I := \{I^0, I^1, ..., I^{L-1}\}$, $F := \{F^0, F^1, ..., F^{L-1}\}$, and $O := \{O^0, O^1, ..., O^{L-1}\}$, respectively, where $L$ is the number of total frames.

\subsection{Network Architecture}
Following existing two-stream approaches for unsupervised VOS, such as MATNet~\cite{MATNet}, FSNet~\cite{FSNet}, HFAN~\cite{HFAN}, PMN~\cite{PMN}, and TMO~\cite{TMO}, our network is designed based on a simple two-stream encoder-decoder architecture. As both RGB image and optical flow map are given as input, two separate encoders are adopted. The features obtained from those encoders are decoded using a decoder that outputs a binary segmentation mask. In the middle of the encoding and decoding processes, the proposed IMA and IFA are adopted for mutual modality fusion and temporal cue aggregation, respectively. The visualized pipeline of our network can be found at Figure~\ref{figure2}.

\subsection{Inter-Modality Attention (IMA)}
\label{IMA}
\begin{figure}[t]
\centering
\includegraphics[width=1\linewidth]{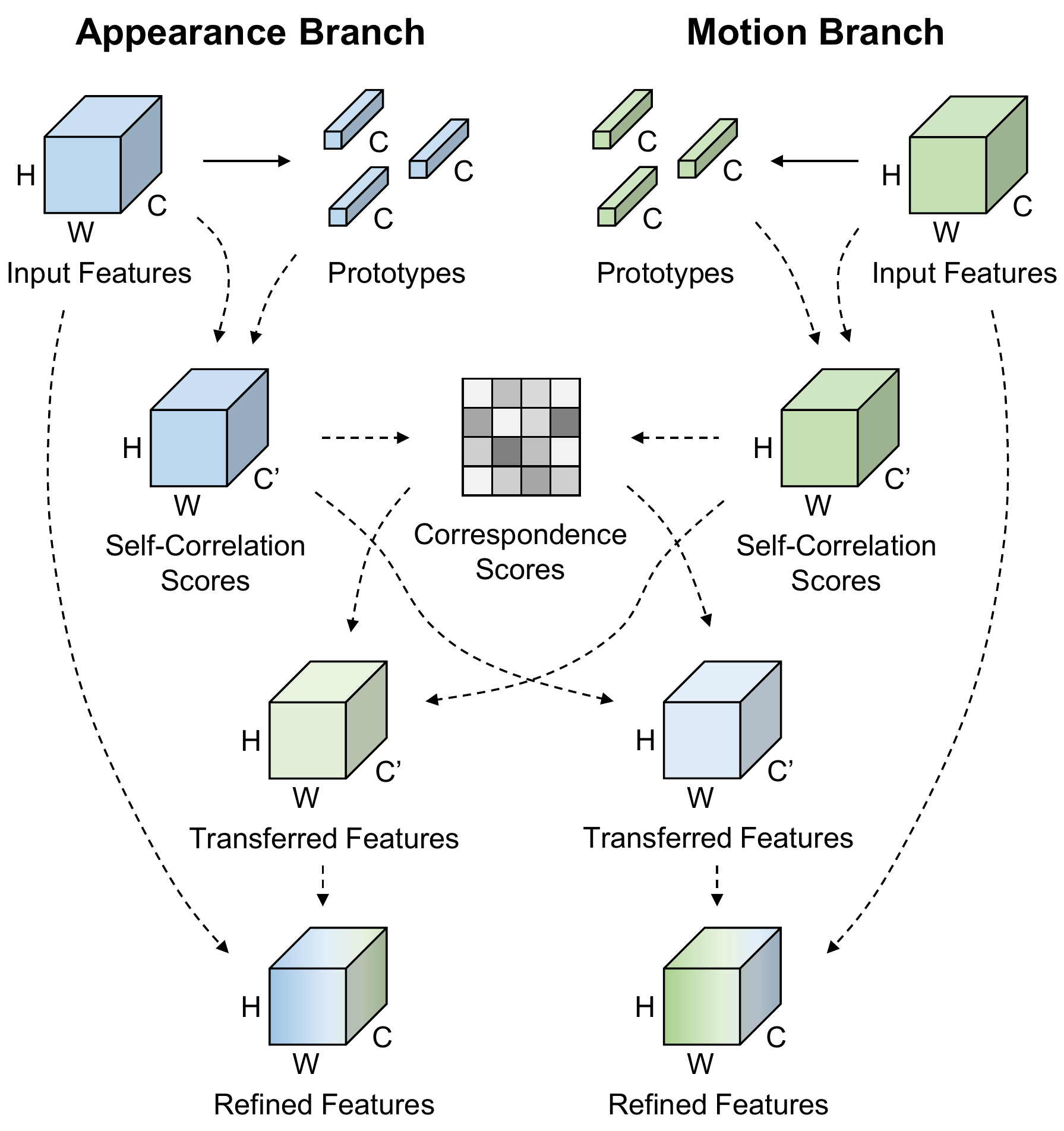}
\caption{Visualized pipeline of IMA.}
\label{figure3}
\end{figure}

Existing two-stream approaches, including the aforementioned methods, focus on fusing the multi-modal cues, i.e., appearance cues and motion cues. However, the fusion has been implemented using a simple summation or concatenation, resulting in unstable cue generation, particularly in challenging scenarios. To enhance multi-modality fusion for unsupervised VOS, we propose IMA to densely and thoroughly exchange information between appearance and motion cues based on prototype attention mechanism. The proposed IMA consists of three parts: prototype generation, self-correlation calculation, and mutual feature refinement. In Figure~\ref{figure3}, we visualize the architecture of IMA.

\vspace{1mm}
\noindent\textbf{Prototype generation.} Inspired by OCR~\cite{OCR}, we first generate prototypes based on learnable object regions. For each input feature map $X \in \mathbb{R}^{C\times HW}$, soft object region $S \in [0, 1]^{C\times HW}$ is calculated by applying a simple channel-wise softmax operation as
\begin{align}
&S = Softmax(X)~.
\end{align}
Considering the backbone encoder is learned with large-scale ImageNet~\cite{imagenet}, each channel in $S$ already contains the clustering ability that helps spatially separate the input feature map into semantic parts. Then, using $X$ and $S$, prototypes $P1 \in \mathbb{R}^{C \times C'}$ are obtained as
\begin{align}
&P1 = X \otimes S^{T}~,
\end{align}
where $\otimes$ indicates matrix multiplication. In $P1$, $C'$ prototypes with a channel size of $C$ are contained. Note that $C'$ is equal to $C$, but is adopted for better clarification.

\vspace{1mm}
\noindent\textbf{Self-correlation calculation.} In order to incorporate the information of the constructed prototypes $P1$ into the input features $X$, we first calculate self-correlation map $\Psi1 \in [-1, 1]^{C'\times HW}$ for each modality as
\begin{align}
&\Psi1 = \mathcal{N}(P1)^{T} \otimes \mathcal{N}(X)~,
\end{align}
where $\mathcal{N}$ indicates channel L2 normalization. The generated $\Psi1$ represents the cosine similarities between each prototype and each input feature. Here, we embed key and value features of each modality from $\Psi1$ instead of directly extracting them from input features $X$. To validate the incorporation of a prototype framework helps better fusion of two modalities, we conduct an ablation study regarding the key-value extraction (normal embedding vs. prototype-based self-correlation calculation) in Section~\ref{analysis}.

\vspace{1mm}
\noindent\textbf{Mutual feature refinement.} After generating the self-correlation maps for each modality, the information of each modality can now be effectively transferred to each other using a cross attention mechanism. From $\Psi1$, the key features $K \in \mathbb{R}^{C' \times HW}$ and value features $V \in \mathbb{R}^{C' \times HW}$ are first calculated as
\begin{align}
&K = \sigma_K(\Psi1)\nonumber\\
&V = \sigma_V(\Psi1)~,
\end{align}
where $\sigma$ indicates a pixel-wise (HW-wise) fully connected layer. Next, the correspondence map $\Phi1 \in \mathbb{R}^{C' \times C'}$ of the key features from each modality are computed as
\begin{align}
&\Phi1 = K_A \otimes K_M^{T}~.
\end{align}
As each channel in $K$ contains certain properties of prototypes, relations between every constructed prototype are contained in $\Phi1$. Based on $\Phi1$, $V$ from one modality are propagated to the other modality as
\begin{align}
&T_A = Softmax(\Phi1) \otimes V_M\nonumber\\
&T_M = Softmax(\Phi1^{T}) \otimes V_A~,
\end{align}
where $T$ indicates the transferred features. Finally, the input features are concatenated with the transferred features, and the refined features $X' \in \mathbb{R}^{C \times HW}$ are obtained as
\begin{align}
&X_A' = Conv(X_A \oplus T_A)\nonumber\\
&X_M' = Conv(X_M \oplus T_M)~,
\end{align}
where $\oplus$ indicates channel concatenation and $Conv$ is a convolutional layer for feature refinement. Through this mutual refinement process, each modality can appropriately reflect semantic information from another modality.

\subsection{Inter-Frame Attention (IFA)}
\begin{figure}[t]
\centering
\includegraphics[width=1\linewidth]{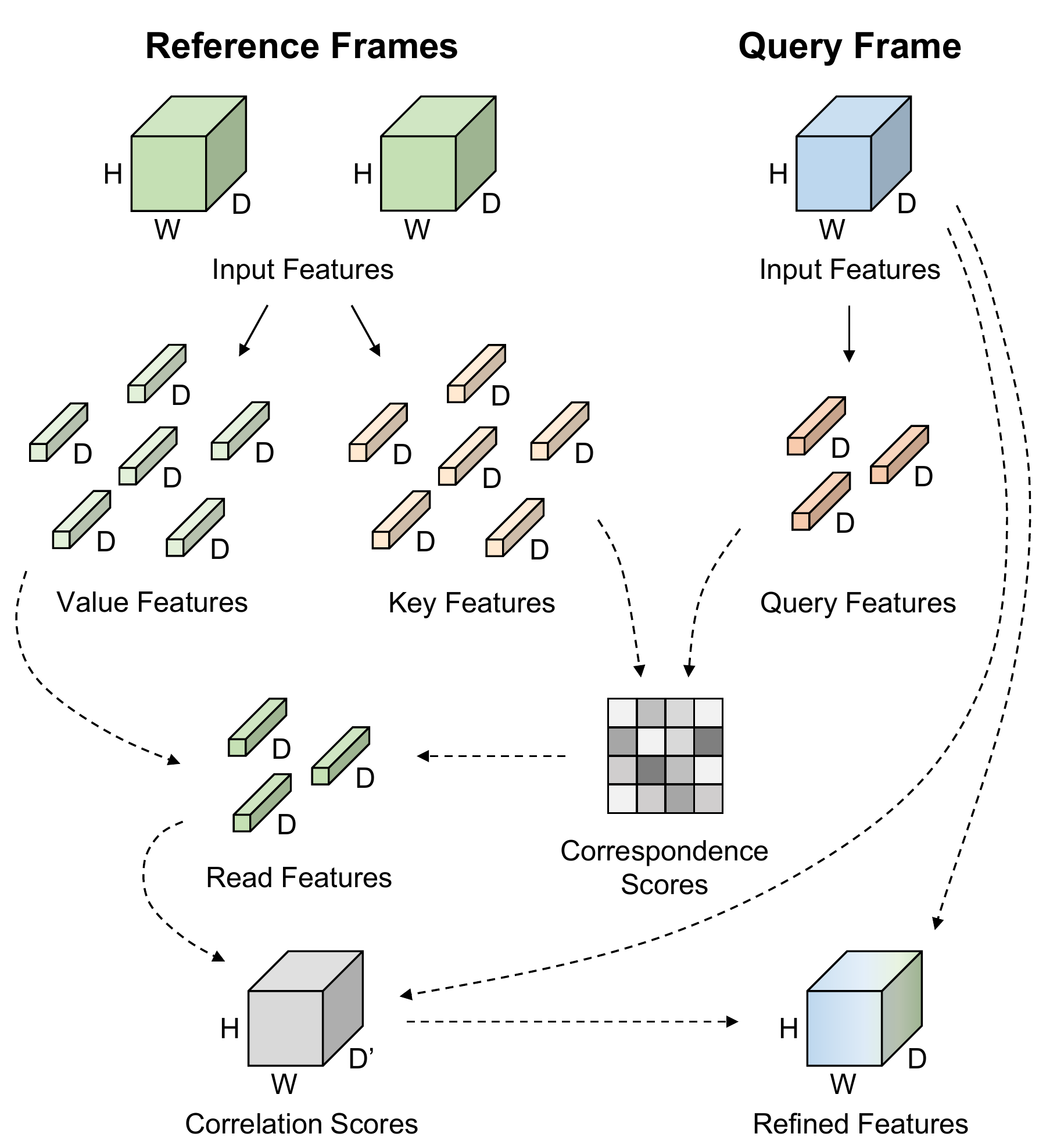}
\caption{Visualized pipeline of IFA.}
\label{figure4}
\end{figure}

As much as fusing the features of multiple modalities is important, exploiting temporal coherence of a video is also an effective strategy for unsupervised VOS. Existing approaches, such as COSNet~\cite{COSNet}, AD-Net~\cite{AD-Net}, F2Net~\cite{F2Net}, and IMP~\cite{IMP}, design their network architectures to utilize this temporal coherence. However, they are either time-consuming owing to their iterative workflows or not fully leveraging the global context of a video. To overcome these limitations, we propose IFA to efficiently leverage the temporal coherence of a video. The proposed IFA consists of three parts: reference frame sampling, prototype generation, and temporal propagation. In Figure~\ref{figure4}, we visualize the architecture of IFA.

\vspace{1mm}
\noindent\textbf{Reference frame sampling.} The objective of IFA is to collect and store the global context of a video and propagate it to each query frame. To efficiently obtain the global context, we sample the frames in a video and store only the sampled frames rather than storing all frames. As the sampling method, we adopt uniform sampling strategy, i.e., frames are sampled while keeping the intervals identical. For example, if we want to sample $N$ frames from $L$-length video, the sampled frames are defined as $\{I^k\}_{i=0}^{N-1}$ where
\begin{align}
&k = \lfloor\frac{i * (L - 1)}{N - 1}\rfloor~.
\end{align}

\vspace{1mm}
\noindent\textbf{Prototype generation.} Before transferring semantic context of the reference frames to the query frame, we first transform the input features $Y \in \mathbb{R}^{D \times HW}$ to prototypes $P2 \in \mathbb{R}^{D \times D'}$ similar to \textit{prototype generation} in Section~\ref{IMA}. $Y$ can be directly used instead of $P2$ for attention embedding in later step, but we use $P2$ to obtain better feature representations. The effects of this feature-prototype conversion process is reported in Section~\ref{analysis}.

\vspace{1mm}
\noindent\textbf{Temporal context propagation.} After constructing prototypes, we extract key features $K \in \mathbb{R}^{D \times ND'}$ and value features $V \in \mathbb{R}^{D \times ND'}$ from the reference frames' prototypes, and query features $Q \in \mathbb{R}^{D \times D'}$ from the query frame's prototypes. Here, all embedding processes are implemented for each frame separately. The correspondence map $\Phi2 \in \mathbb{R}^{D' \times ND'}$ can be obtained as
\begin{align}
&\Phi2 = {Q}^{T} \otimes K~.
\end{align}
Based on $\Phi2$, the context of the reference frames is adaptively read and stored in read features $R \in \mathbb{R}^{D \times D'}$ as
\begin{align}
&R = (Softmax(\Phi2) \otimes {V}^{T})^{T}~.
\end{align}
The generated $R$ has $D'$ prototypes with feature size of $D$, which contain information of the sampled frames. As it does not have spatial information related to the query frame, it cannot be directly leveraged for the feature fusion process. Therefore, we calculate the correlation scores $\Psi2 \in [-1, 1]^{D'\times HW}$ between the query frame's $Y$ and $R$ to force the temporally transferred information to have the same spatial size as the input features, as follows:
\begin{align}
&\Psi2 = \mathcal{N}(R)^{T} \otimes \mathcal{N}(Y)~.
\end{align}
As $Y$ and $R$ have the same spatial size, feature fusion between them can now be easily achieved. The refined features $Y' \in \mathbb{R}^{D \times HW}$ for the query frame is defined as
\begin{align}
&Y' = Conv(Y \oplus \Psi2)~.
\end{align}
By employing the proposed IFA, semantic context of the reference frames is propagated to the query frame for reliable cue generation. In particular, when reliable information cannot be obtained from a single frame owing to difficulties such as occlusions, the use of IFA can effectively lead to stable functioning of a system.

\subsection{Implementation Details}
\noindent\textbf{Optical flow map.} Following a common protocol for two-stream approaches in unsupervised VOS, we generate optical flow maps from a pre-trained optical flow estimation model. The generated two-channel motion flow maps are converted to three-channel RGB flow maps and then saved in advance. As an optical flow estimation model, we adopt RAFT~\cite{RAFT} pre-trained on the Sintel~\cite{Sintel} dataset.

\vspace{1mm}
\noindent\textbf{Network design.} We adopt VGG-16~\cite{vgg} as our backbone encoder for the appearance branch and motion branch. The encoded features are refined using IMA and IFA, where IMA is adopted for the fourth and fifth encoding blocks and IFA is adopted at the fifth encoding block. Note that IMA and IFA are separately adopted in the fifth encoding block, that is, they are employed in a parallel manner and then fused later. After refining the encoded features using IMA and IFA, an ASPP module~\cite{ASPP} is applied to obtain stronger feature representations. The decoder takes the features from the ASPP module as its input, and gradually refines those features using lower-level features from the encoder.

\vspace{1mm}
\noindent\textbf{Two-stage network training.} Following previous methods, such as F2Net~\cite{F2Net}, RTNet~\cite{RTNet}, FSNet~\cite{FSNet}, and PMN~\cite{PMN}, we train our network using multiple steps. As the first step, a salient object detection dataset DUTS~\cite{DUTS} is adopted to pre-train the model on large-scale data. Both DUTS training set and test set are used as our training dataset. As it is an image-level dataset, only RGB images are available. Therefore, we only train the appearance branch and copy the learned parameters to the motion branch after the pre-training is done. Then, the entire model is trained on the DAVIS 2016~\cite{DAVIS} training set and YouTube-VOS 2018~\cite{YTVOS} training set with both appearance branch and motion branch. If a video sequence contains multiple objects, we regard them as a single object to obtain binary ground truth masks. Training snippets are randomly sampled from the DAVIS 2016 training set and YouTube-VOS 2018 training set with the same probabilities. The length of each snippet is fixed at four frames.

\vspace{1mm}
\noindent\textbf{Training details.} For network optimization, we use cross-entropy loss and the Adam optimizer~\cite{adam}. The learning rate is decayed from 1e-4 to 1e-5 using the cosine annealing scheduler~\cite{cosine}, and the batch size is set to 16. For network training, two GeForce RTX 3090 GPUs are used.

\section{Experiments}
In Section~\ref{datasets} and Section~\ref{metrics}, the datasets and metrics used in this study are first introduced. Each proposed component is analyzed in Section~\ref{analysis}. Quantitative and qualitative comparison can be found at Section~\ref{quanti} and Section~\ref{quali}, respectively. Our method is abbreviated as DPA.

\subsection{Datasets}
\label{datasets}
In this study, we use three datasets for network training: DUTS~\cite{DUTS} dataset, DAVIS 2016~\cite{DAVIS} training set, and YouTube-VOS 2018~\cite{YTVOS} training set; and three datasets for network testing: DAVIS 2016 validation set, FBMS~\cite{FBMS} test set, and YouTube-Objects~\cite{YTOBJ} dataset.

\subsection{Evaluation Metrics}
\label{metrics}
We employ three evaluation metrics in this study: region similarity $\mathcal{J}$, boundary accuracy $\mathcal{F}$, and their average $\mathcal{G}$. $\mathcal{J}$ and $\mathcal{F}$ can be calculated as follows:
\begin{align}
&\mathcal{J} = \left| \frac{M_{gt} \cap M_{pred}}{M_{gt} \cup M_{pred}} \right|~,
\label{eq3}
\end{align}
\begin{align}
&\mathcal{F} = \frac{2 \times \text{Precision} \times \text{Recall}}{\text{Precision} + \text{Recall}}~.
\end{align}
For the DAVIS 2016~\cite{DAVIS} validation set, $\mathcal{G}$, $\mathcal{J}$, and $\mathcal{F}$ are used for network evaluation, while only $\mathcal{J}$ is used for the FBMS~\cite{FBMS} test set and YouTube-Objects~\cite{YTOBJ} dataset.

\begin{table}[t!]
\centering 
\caption{Ablation study on the proposed components. $P$ and $N$ indicate the use of prototype embedding and the number of reference frames used in IFA, respectively.}
\vspace{-2mm}
\small
\begin{tabular}{c|c|c|P{0.6cm}|P{0.6cm}P{0.6cm}P{0.6cm}}
\toprule
Version &IMA &IFA &$N$ &$\mathcal{G}_\mathcal{M}$ &$\mathcal{J}_\mathcal{M}$ &$\mathcal{F}_\mathcal{M}$\\
\midrule
\Romannum{1} &$\times$ &$\times$ &- &83.4 &83.2 &83.5\\
\Romannum{2} &w/ $P$ &$\times$ &- &85.9 &85.4 &86.3\\
\Romannum{3} &$\times$ &w/ $P$ &4 &85.4 &85.0 &85.8\\
\Romannum{4} &w/ $P$ &w/ $P$ &4 &86.9 &86.3 &87.4\\
\midrule
\Romannum{5} &w/ $P$ &w/ $P$ &1 &86.3 &85.8 &86.9\\
\Romannum{6} &w/ $P$ &w/ $P$ &2 &86.5 &86.0 &87.1\\
\Romannum{7} &w/ $P$ &w/ $P$ &3 &86.8 &86.2 &87.5\\
\Romannum{8} &w/ $P$ &w/ $P$ &5 &86.9 &86.3 &87.5\\
\midrule
\Romannum{9} &w/o $P$ &w/ $P$ &4 &86.1 &85.5 &86.6\\
\Romannum{10} &w/ $P$ &w/o $P$ &4 &86.3 &85.7 &87.1\\
\Romannum{11} &w/o $P$ &w/o $P$ &4 &85.3 &84.6 &86.0\\
\bottomrule
\end{tabular}
\label{table1}
\end{table}

\begin{figure}[t]
\centering
\includegraphics[width=1\linewidth]{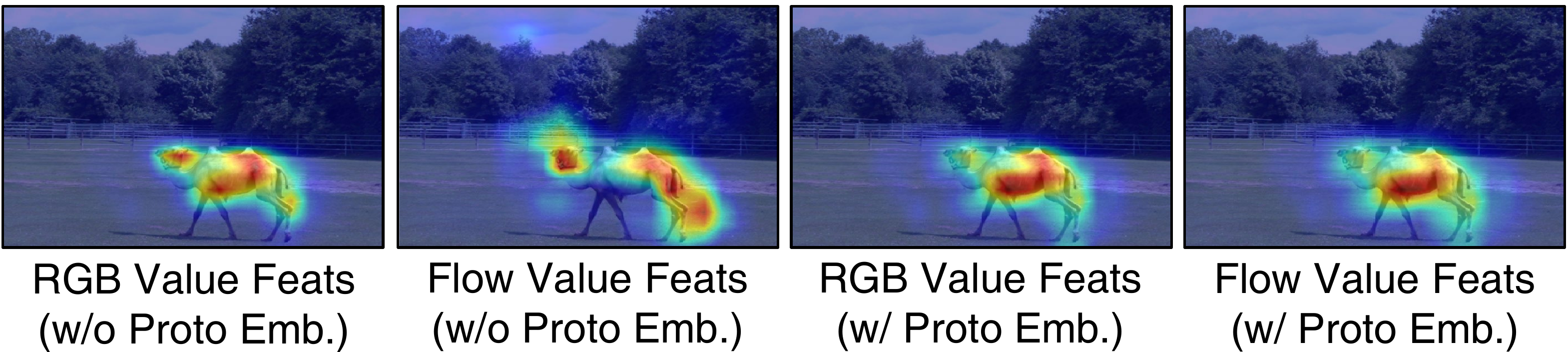}
\caption{Visualized activation maps of different IMA versions.}
\label{figure5}
\end{figure}

\subsection{Analysis}
\label{analysis}
To verify the effectiveness of the proposed components, we perform an ablation study on them, the results of which are present in Table~\ref{table1} and Table~\ref{table2}. Note that the models are trained and tested with 352$\times$352-resolution videos and evaluated on the DAVIS 2016 validation set.

\noindent\textbf{Use of IMA and IFA.} To compare the model performance with and without IMA and IFA, we compare model \Romannum{1}, \Romannum{2}, \Romannum{3}, and \Romannum{4} in Table~\ref{table1}. As presented in the table, IMA and IFA both bring significant performance improvements to the baseline model. When IMA is employed alone, 2.5\% improvements on $\mathcal{G}_\mathcal{M}$ are obtained, which implies that identifying inter-relationships between RGB images and optical flow maps via cross attention is effective for blending two distinct streams. IFA also brings meaningful improvements, $\mathcal{G}_\mathcal{M}$ score of 2.0\%, backing up the need for global observation for specifying the primary objects. If IMA and IFA are used together, outstanding performance is obtained, $\mathcal{G}_\mathcal{M}$ of 86.9\%. This indicates that IMA and IFA can constructively compensate each other as they focus on separate problems lying in two-stream unsupervised VOS.

Qualitative effects of adopting IMA and IFA can also be found at Figure~\ref{figure1}. In the figure, feature maps of various embedding stages are visualized. We compare four feature maps, i.e., feature maps before IMA and IFA, feature maps after IMA, feature maps after IFA, and feature maps after IMA and IFA. It can be seen that IMA and IFA are both effective for capturing and specifying the salient objects (clearer edges and higher confidence for object regions). The joint use of IMA and IFA is even better than using IMA or IFA alone, validating the compatibility.

\vspace{1mm}
\noindent\textbf{Number of reference frames.} As described, IFA can take an arbitrary number of frames as reference frames given that it is based on an attention mechanism. To determine the optimal number of frames, we compare models with different number of reference frames. As shown in model \Romannum{5}, \Romannum{6}, \Romannum{7}, and \Romannum{8} in Table~\ref{table1}, employing more reference frames generally leads to a higher segmentation performance. This proves that as the amount of information from a video increases, the model becomes more generalized and robust against challenges such as occlusion. However, if the number of reference frames is larger than three, the performance gain is not satisfactory considering additional computations. In other words, semantic cues obtained from three representative frames are generally sufficient to encompass the global properties of a video.

\vspace{1mm}
\noindent\textbf{Prototype embedding.} In IMA and IFA, we employ a prototype framework before the attention mechanism as a pre-processing step. In Table~\ref{table1}, we quantitatively compare the model versions with and without this protocol. For both IMA and IFA, it brings meaningful improvements, demonstrating its effectiveness as a feature refinement tool. In Figure~\ref{figure5}, we also visually compare the value feature maps with input features and self-correlation scores as the embedding source. As discussed in TMO~\cite{TMO}, when there is a difference in quality between two domains, the information from the less reliable domain can act as noise. Considering this, applying prototype embedding significantly increases the stability of the network, as the features from both domains focus on the same regions. The qualitative results align with improvements in quantitative performance.

\vspace{1mm}
\noindent\textbf{Cost analysis.} In Table~\ref{table2}, we compare the number of parameters and inference time of different model versions. As shown in the table, IMA and IFA only introduce a small increase in the number of parameters compared to the baseline model. While there is some inference time slowdown due to the incremental computational cost, the performance gain is substantial considering the trade-offs.

\begin{table}[t!]
\centering 
\caption{Cost analysis of the proposed components.}
\vspace{-2mm}
\small
\begin{tabular}{c|P{0.7cm}|P{0.7cm}|ccc}
\toprule
Version &IMA &IFA &Param \# &Time (s) &$\mathcal{G}_\mathcal{M}$\\
\midrule
\Romannum{1} & & &39.5M &0.0164 &83.4\\
\Romannum{2} &\checkmark & &41.5M &0.0183 &85.9\\
\Romannum{3} & &\checkmark &40.5M &0.0322 &85.4\\
\Romannum{4} &\checkmark &\checkmark &43.5M &0.0414 &86.9\\
\bottomrule
\end{tabular}
\label{table2}
\end{table}

\begin{table*}
\centering 
\caption{Quantitative evaluation on the DAVIS 2016 validation set and FBMS test set. OF and PP indicate the use of optical flow estimation models and post-processing techniques, respectively. * denotes speed calculated on our hardware.}
\vspace{-2mm}
\small
\begin{tabular}{p{2.3cm}P{2cm}P{1.5cm}P{0.5cm}P{0.5cm}P{1cm}P{1cm}P{1cm}P{1cm}P{1cm}}
\toprule
\multicolumn{6}{c}{} &\multicolumn{3}{c}{DAVIS 2016} &\multicolumn{1}{c}{FBMS}\\
\cline{7-10}
Method &Publication &Resolution &OF &PP &fps &$\mathcal{G}_\mathcal{M}$ &$\mathcal{J}_\mathcal{M}$ &$\mathcal{F}_\mathcal{M}$ &$\mathcal{J}_\mathcal{M}$\\
\midrule
PDB~\cite{PDB} &ECCV'18 &473$\times$473 & &\checkmark &20.0 &75.9 &77.2 &74.5 &74.0\\
MOTAdapt~\cite{MOTAdapt} &ICRA'19 &- & &\checkmark &- &77.3 &77.2 &77.4 &-\\
AGS~\cite{AGS} &CVPR'19 &473$\times$473 & &\checkmark &10.0 &78.6 &79.7 &77.4 &-\\
COSNet~\cite{COSNet} &CVPR'19 &473$\times$473 & &\checkmark &- &80.0 &80.5 &79.4 &75.6\\
AD-Net~\cite{AD-Net} &ICCV'19 &480$\times$854 & &\checkmark &4.00 &81.1 &81.7 &80.5 &-\\
AGNN~\cite{AGNN} &ICCV'19 &473$\times$473 & &\checkmark &3.57 &79.9 &80.7 &79.1 &-\\
MATNet~\cite{MATNet} &AAAI'20 &473$\times$473 &\checkmark &\checkmark &20.0 &81.6 &82.4 &80.7 &76.1\\
WCS-Net~\cite{WCS-Net} &ECCV'20 &320$\times$320 & & &33.3 &81.5 &82.2 &80.7 &-\\
DFNet~\cite{DFNet} &ECCV'20 &- & &\checkmark &3.57 &82.6 &83.4 &81.8 &-\\
3DC-Seg~\cite{3DC-Seg} &BMVC'20 &480$\times$854 & &\checkmark &4.55 &84.5 &84.3 &84.7 &-\\
F2Net~\cite{F2Net} &AAAI'21 &473$\times$473 & & &10.0 &83.7 &83.1 &84.4 &77.5\\
RTNet~\cite{RTNet} &CVPR'21 &384$\times$672 &\checkmark &\checkmark &- &85.2 &85.6 &84.7 &-\\
FSNet~\cite{FSNet} &ICCV'21 &352$\times$352 &\checkmark &\checkmark &12.5 &83.3 &83.4 &83.1 &-\\
TransportNet~\cite{TransportNet} &ICCV'21 &512$\times$512 &\checkmark & &12.5 &84.8 &84.5 &85.0 &78.7\\
AMC-Net~\cite{AMC-Net} &ICCV'21 &384$\times$384 &\checkmark &\checkmark &17.5 &84.6 &84.5 &84.6 &76.5\\
D$^2$Conv3D~\cite{D^2Conv3D} &WACV'22 &480$\times$854 & & &- &86.0 &85.5 &86.5 &-\\
IMP~\cite{IMP} &AAAI'22 &- & & &1.79 &85.6 &84.5 &86.7 &77.5\\
HFAN~\cite{HFAN} &ECCV'22 &512$\times$512 &\checkmark & &11.0$^*$ &\underline{87.0} &\underline{86.6} &87.3 &-\\
PMN~\cite{PMN} &WACV'23 &352$\times$352 &\checkmark & &\underline{41.3}$^*$ &85.9 &85.4 &86.4 &77.7\\
TMO~\cite{TMO} &WACV'23 &384$\times$384 &\checkmark & &\textbf{43.2}$^*$ &86.1 &85.6 &86.6 &79.9\\
OAST~\cite{OAST} &ICCV'23 &384$\times$640 &\checkmark & &- &85.9 &85.4 &86.3 &\underline{81.9}\\
\midrule 
\textbf{DPA} & &352$\times$352 &\checkmark & &24.2$^*$ &86.9 &86.3 &\underline{87.4} &81.2\\
\textbf{DPA} & &512$\times$512 &\checkmark & &19.5$^*$ &\textbf{87.6} &\textbf{86.8} &\textbf{88.4} &\textbf{83.4}\\
\bottomrule
\end{tabular}
\label{table3}
\end{table*}

\begin{table*}
\centering 
\caption{Quantitative evaluation on the YouTube-Objects dataset. Performance is reported using the $\mathcal{J}$ mean.}
\vspace{-2mm}
\small
\begin{tabular}{p{1.9cm}P{1.2cm}P{0.9cm}P{0.9cm}P{0.9cm}P{0.9cm}P{0.9cm}P{0.9cm}P{0.9cm}P{1.2cm}P{0.9cm}|P{0.9cm}}
\toprule
Method &Aeroplane &Bird &Boat &Car &Cat &Cow &Dog &Horse &Motorbike &Train &Mean\\
\midrule
AGS~\cite{AGS} &\textbf{87.7} &76.7 &\textbf{72.2} &78.6 &69.2 &64.6 &73.3 &64.4 &62.1 &48.2 &69.7\\
COSNet~\cite{COSNet} &81.1 &75.7 &\underline{71.3} &77.6 &66.5 &69.8 &76.8 &\underline{67.4} &67.7 &46.8 &70.5\\
AGNN~\cite{AGNN} &71.1 &75.9 &70.7 &78.1 &67.9 &69.7 &\underline{77.4} &67.3 &\underline{68.3} &47.8 &70.8\\
MATNet~\cite{MATNet} &72.9 &77.5 &66.9 &79.0 &\underline{73.7} &67.4 &75.9 &63.2 &62.6 &51.0 &69.0\\
WCS-Net~\cite{WCS-Net} &81.8 &\underline{81.1} &67.7 &79.2 &64.7 &65.8 &73.4 &\textbf{68.6} &\textbf{69.7} &49.2 &70.5\\
RTNet~\cite{RTNet} &84.1 &80.2 &70.1 &\underline{79.5} &71.8 &\underline{70.1} &71.3 &65.1 &64.6 &\underline{53.3} &71.0\\
AMC-Net~\cite{AMC-Net} &78.9 &80.9 &67.4 &\textbf{82.0} &69.0 &69.6 &75.8 &63.0 &63.4 &\textbf{57.8} &71.1\\
TMO~\cite{TMO} &85.7 &80.0 &70.1 &78.0 &73.6 &\textbf{70.3} &76.8 &66.2 &58.6 &47.0 &\underline{71.5}\\
\midrule
\textbf{DPA} &\underline{87.5} &\textbf{85.6} &70.1 &77.7 &\textbf{81.2} &69.0 &\textbf{81.8} &61.9 &62.1 &51.3 &\textbf{73.7}\\
\bottomrule
\end{tabular}
\label{table4}
\end{table*}

\subsection{Quantitative Results}
\label{quanti}
In Table~\ref{table3} and Table~\ref{table4}, we present quantitative comparison between our proposed method and existing state-of-the-art methods on the DAVIS 2016~\cite{DAVIS} validation set, FBMS~\cite{FBMS} test set, and YouTube-Objects~\cite{YTOBJ} dataset. Note that all methods are evaluated using CNN-based backbone networks for a fair comparison. Our model is tested on a single GeForce RTX 2080 Ti GPU.

\vspace{1mm}
\noindent\textbf{DAVIS 2016.} On the DAVIS 2016 validation set, the methods using the estimated optical flow maps generally exhibit notable performance. Specifically, recent methods achieve high performance based on a two-stream architecture without the need for post-processing steps. Our proposed DPA outperforms all other methods at the same input resolution. With a resolution of 352$\times$352, it achieves a $\mathcal{G}_\mathcal{M}$ score of 86.9\%, whereas with a higher resolution of 512$\times$512, a $\mathcal{G}_\mathcal{M}$ score of 87.6\% is obtained. While demonstrating such satisfactory performance, the inference speed is also comparable to other fast methods.

\begin{figure*}[t]
\centering
\includegraphics[width=1\linewidth]{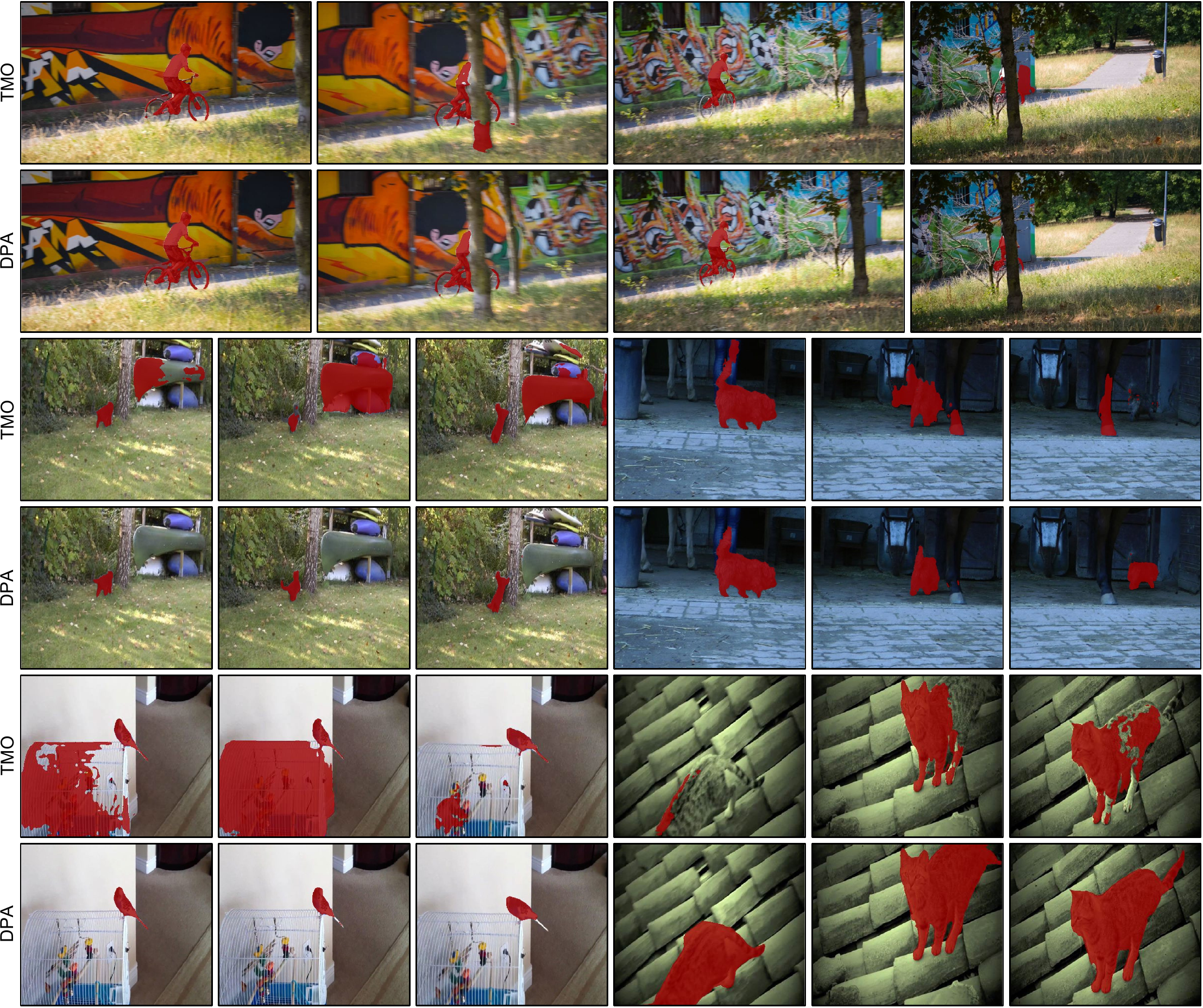}
\caption{Qualitative comparison between state-of-the-art TMO and the proposed DPA.}
\label{figure6}
\end{figure*}

\vspace{1mm}
\noindent\textbf{FBMS.} Unlike the DAVIS 2016 validation set, the FBMS test set also contains multi-object scenarios as well as the single-object scenarios. DPA outperforms all other existing approaches by a significant margin with a $\mathcal{J}_\mathcal{M}$ score of 1.5\%, which demonstrates robustness of DPA even against videos containing multiple objects.

\vspace{1mm}
\noindent\textbf{YouTube-Objects.} Compared to the DAVIS 2016 validation set and FBMS test set, the YouTube-Objects dataset presents a significantly more challenging environment, as salient objects are often less distinctive. DPA surpasses all other methods on the YouTube-Objects dataset, showcasing its effectiveness even in challenging scenarios.

\subsection{Qualitative Results}
\label{quali}
In Figure~\ref{figure6}, we qualitatively compare our proposed DPA to the state-of-the-art TMO~\cite{TMO}. While TMO is often distracted by background elements, DPA consistently captures the primary objects. Even when the target object is occluded by obstacles, DPA maintains stable tracking of the object.

\section{Conclusion}
In unsupervised VOS, multi-modality fusion and temporal aggregation are recognized as essential components. However, they have limitations, such as a lack of thorough information exchange or substantial time consumption. To address these limitations and further enhance performance, we propose two novel attention modules, IMA and IFA. Additionally, each module is further enhanced by incorporating a prototype framework. By leveraging IMA and IFA in a collaborative manner, we achieve outstanding performance on all public benchmark datasets.

\vspace*{\fill}
\noindent\textbf{Acknowledgement.} This work was supported in part by the Korea Institute of Science and Technology (KIST) Institutional Program under Project 2E33001.

{
    \small
    \bibliographystyle{ieeenat_fullname}
    \bibliography{DPA}
}

\end{document}